\documentclass[conference,a4paper]{IEEEtran}
\IEEEoverridecommandlockouts
\usepackage{cite}
\usepackage{amsmath,amssymb,amsfonts}
\usepackage{algorithmic}
\usepackage{graphicx}
\usepackage{textcomp}
\usepackage{xcolor}
\usepackage{balance}
\usepackage[version=4]{mhchem}
\usepackage[utf8]{inputenc}
\usepackage{float}

\usepackage{bigstrut}
\usepackage{multirow}

\usepackage{subcaption}
\usepackage[font=small,labelfont=bf]{caption}
\usepackage[belowskip=-2pt,aboveskip=5pt]{caption}

\def\BibTeX{{\rm B\kern-.05em{\sc i\kern-.025em b}\kern-.08em
    T\kern-.1667em\lower.7ex\hbox{E}\kern-.125emX}}
\begin{document}

\title{Does complimentary information from multispectral imaging improve face presentation attack detection? \\
%Multi-spectral Imaging To Distinguish Natural and Artificial Ripened Banana

}

\author{\IEEEauthorblockN{Narayan Vetrekar$^{\dagger}$ \quad Raghavendra Ramachandra$^{\ddagger}$ \quad Sushma Venkatesh$^{\$}$ \quad Jyoti D. Pawar$^{*}$ \quad R. S. Gad$^{\dagger}$}
\IEEEauthorblockA{\textit{$^{\dagger}$School of Physical and Applied Sciences, Goa University, Goa, India} \\
\textit{$^{\ddagger}$Norwegian University of Science and Technology (NTNU), Gj{\o}vik, Norway}\\
\textit{$^{*}$Computer science and Technology, Goa Business School, Goa University, Goa, India}\\
\textit{$^{\$}$AiBA AS, Gj{\o}vik, Norway}\\
%Goa, India \\
E-mail: \{vetrekarnarayan; jdp; rsgad\} @unigoa.ac.in,
\{raghavendra.ramachandra\} @ntnu.no}
%\and
%\and
%\and
%\and
%\and
}
\maketitle

\begin{abstract}
Presentation Attack Detection (PAD) has been extensively studied, particularly in  the visible spectrum. With the advancement of sensing technology beyond the visible range, multispectral imaging has gained significant attention in this direction. We present PAD based on multispectral images constructed for eight different presentation artifacts resulted from three different artifact species. In this work, we introduce Face Presentation Attack Multispectral (FPAMS) database to demonstrate the significance of employing multispectral imaging. The goal of this work is to study complementary information that can be combined in two different ways (image fusion and score fusion) from multispectral imaging to improve the face PAD.  The experimental evaluation results present an extensive qualitative analysis of $61650$ sample multispectral images collected for bonafide and artifacts.  The PAD based on the score fusion and image fusion method presents superior performance, demonstrating the significance of employing multispectral imaging to detect presentation artifacts.     
\end{abstract}


\section{Introduction}
\label{sec:introduction}

\begin{figure}[t!]
	\centering
	\includegraphics[width = 1\linewidth]{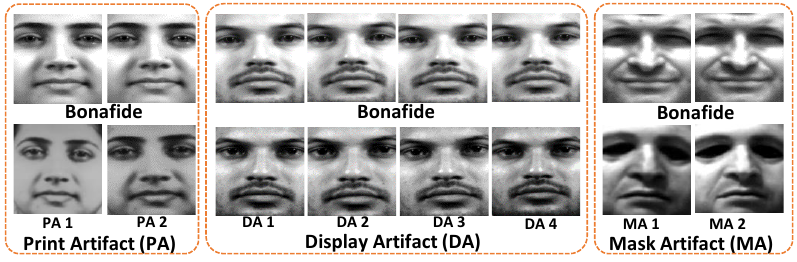}
	\caption{Sample images illustrates the variation in different facial artifacts. Top row - Bonafide samples, Bottom left - Print Artifact, Bottom middle - Display Artifacts and Bottom right - Mask Artifacts}
	\label{fig:PAD-samples}
\end{figure}

Over the past decades, the verification of users based on physiological biometric modalities has been the major reason for the popularity of biometrics for numerous applications\cite{article}. Among the physiological biometric modalities, person verification using face is considered more convenient because of its ease of use and the non-intrusive nature of image acquisition. Despite impressive verification performance, and even outperforming human performance on most challenging datasets, face recognition systems still pose serious challenges when it comes to presentation attacks (PA) (i.e., spoofing attacks) \cite{8714076}. A presentation attack is a deliberate attempt at impostor artifacts to impersonate the identity of genuine users by using Presentation Attack Instruments (PAIs) (according to the definitions of ISO/IEC 30107 standards \cite{ISO-IEC-JTC-SC37-2013}). With the widespread availability of facial images in the public domain, various PAIs are being created by attackers to obtain unauthorized access by presenting fake artifacts. The PAIs could be simple printed photographs or electronics display artifacts that constitute 2D presentation artifacts, while more sophisticated PAIs are 3D face mask artifacts presented in front of the Face Recognition System (FRS) to avail the access. Figure~\ref{fig:PAD-samples} illustrates the PAIs showing 2D and 3D presentation artifacts.         
%---------------%
%%--------------%
The influence of PAIs such as 2D print, electronic display, and sophisticated 3D face masks has been studied in a substantial manner using state-of-the-art methods to demonstrate the vulnerability of facial biometrics against artifacts\cite{8014830} \cite{george2022comprehensive,article,Bhattacharjee2019}. Therefore, to mitigate vulnerability issues, several Presentation Attack Detection (PAD) algorithms based on handcrafted features and deep learning-based approaches have been proposed in the literature \cite{article}. Although we note that the surveillance system operates in the visible spectrum, the majority of the face PADs employed are based on the visible spectrum \cite{george2022comprehensive}. On the other hand, artifacts are non-skin materials that leverage differential illumination properties compared to genuine skin across the electromagnetic spectrum because of which previous work has also shown preferences in working beyond the visible spectrum to alleviate the vulnerability of facial biometric systems \cite{iet:content-journals}. More specifically, multispectral imaging has shown greater potential in this direction, thereby leveraging differential information in spatial and spectral domains. Considering these merits, in our work, we employed a multispectral imaging approach in nine narrow spectrum bands across the Visible (VIS) and Near-Infra-Red (NIR) wavelength ranges to detect presentation artifacts. Furthermore, generalizability towards unseen or unknown artifacts is a challenging task; hence, in this work, we present PAD by exploring the properties of multispectral imaging based on our newly introduced Face Presentation Attack Multispectral Database (FPAMS Database) for unseen or unknown artifacts in order to present the significance of our work. The major contributions of this work are summarized as follows: (1) Present face presentation attack detection explores the inherent properties of multispectral imaging in nine narrow bands across the VIS and NIR ($530nm$ $1000nm$) wavelength range. (2) Quantitative comparison of the image fusion (or early fusion) and score fusion (or late fusion) frameworks for face PAD.  (3) Extensive experimental evaluation results are obtained on the newly introduced FPAMS database of $61650$ samples, especially with the execution protocol of unseen  attack detection, to confirm the performance of the proposed PAD framework. 
%---------------%
\begin{figure*}[htbp!]
	\centering
	\includegraphics[width = 0.9\linewidth]{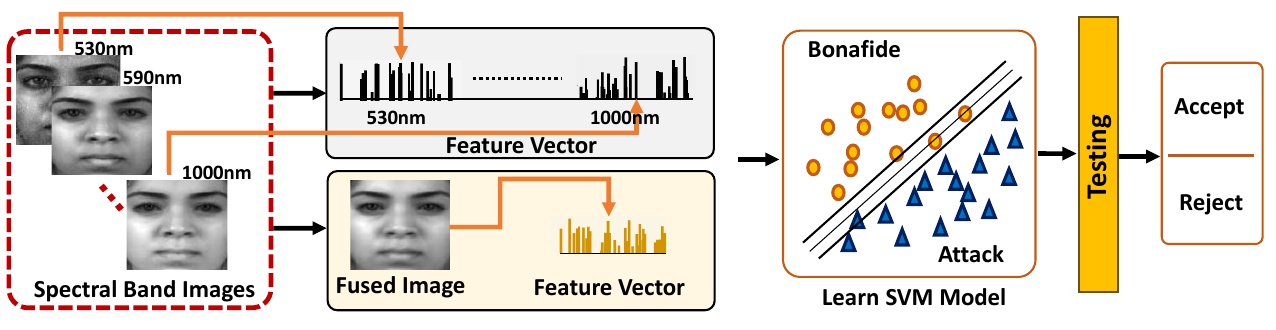}
	\caption{Presentation Attack Detection (PAD) framework}
	\label{fig:PADProposed}
\end{figure*}
%--------------%

%\begin{enumerate}
%	\item Present face presentation attack detection exploring the inherent properties of multi-spectral imaging in nine narrow bands across VIS and NIR wavelength range. Moreover, this study present significance of employing spectral signature, texture descriptor and fusion based approach in PADs essentially for unseen artifacts   
%	\item Extensive experimental evaluations results are performed on newly introduced FPAMS database, which will be made available freely to the researchers.    
%\end{enumerate}       
The rest of the paper is organized as follows: Section~\ref{sec:database} presents a detailed description of the FPAMS database employed in this study, and Section~\ref{sec:method} details the PADs based on image fusion and score fusion  algorithms.  Section~\ref{sec:results} presents the experimental results, and final conclusion is summarized in Section~\ref{sec:conclusion}.

\section{Face Presentation Attack Multi-spectral Database (FPAMS Database)}
\label{sec:database}
FPAMS databases are acquired using custom-built multispectral sensors in nine narrow bands that includes $530nm$, $590nm$, $650nm$, $710nm$, $770nm$, $830nm$, $890nm$, $950nm$, $1000nm$ spanning across the VIS and NIR wavelength range\cite{7738245}. The FPAMS comprises $bonafide$ and $presentation$ $attacks$ acquired under controlled environmental conditions (Refer Figure\ref{fig:PAD-samples}). Further, the $bonafide$ samples were collected in $two$ different sessions separated by a time gap of $three$ to $four$ weeks, whereas the samples associated with $presentation$ $attack$ were acquired in a single session. 
The details of each category of database is briefly presented in the following subsections. %Detail summary of number of samples collected for $Bonafide$ and $Presentation$ $attacks$ are illustrated in Table~\ref{tab:database} and the 
%Figure~\ref{fig:PAD-samples} illustrates the samples spectral band images of FPAMS database. 

%%---------------%
%\begin{figure}[htbp!]
%	\centering
%	\includegraphics[width = 1\linewidth]{PA-database.pdf}
%	\caption{Sample multi-spectral band images collected across nine bands for bonafide and eight presentation artifact}
%	\label{fig:PA-database}
%\end{figure}
%%--------------%
  
\subsection{$Bonafide$ Subset of FPAMS Database}
\label{ssec:bonafide-data}
$Bonafide$ samples of the FPAMS database consisted of sample images collected from $145$ subjects, including $87$ male and $58$ female samples acquired in a control indoor environment. For each session, $5$ sample images were collected and a total of $13050$ samples, which corresponds to $145$ subjects $\times$ $2$ sessions $\times$ $5$ samples $\times$ $9$ bands = $13050$ samples.% to constitute $bonafide$ samples \cite{8311455}. %For more detailed description of $Bonafide$ subset can be obtained from \cite{8311455}.  
\begin{table}[htbp]
	\centering
	\caption{Notations used for different presentation artifact species}
	\resizebox{0.48\textwidth}{!}{
	 % Table generated by Excel2LaTeX from sheet 'Sheet2'
	\begin{tabular}{|c|c|l|}
		\hline
		\textbf{PAIs} & \textbf{Notation} & \textbf{Description} \bigstrut\\
		\hline
		\multicolumn{1}{|c|}{\multirow{2}[4]{*}{Print Artifact (PA)}} & Print Artifact 1 & Laser Printer  \bigstrut\\
		\cline{2-3}       & Print Artifact 2 & Inkjet Printer \bigstrut\\
		\hline \hline
		\multicolumn{1}{|c|}{\multirow{4}[8]{*}{Display Artifact (DA) }} & Display Artifact 1 & Apple iMAC 24-inch 5K Retina Display \bigstrut\\
		\cline{2-3}       & Display Artifact 2 & Dell 27-inch 5K LED Display \bigstrut\\
		\cline{2-3}       & Display Artifact 3 & Apple iPAD 9.7-inch Retina Display \bigstrut\\
		\cline{2-3}       & Display Artifact 4 & Samsung Galaxy S8 58-inch Smartphone \bigstrut\\
		\hline \hline
		\multicolumn{1}{|c|}{\multirow{2}[4]{*}{Mask Artifact (MA)}} & Mask Artifact 1 & Rigid Color 3D Mask \bigstrut\\
		\cline{2-3}       & Mask Artifact 2 & Rigid White 3D Mask \bigstrut\\
		\hline
	\end{tabular}%
	\label{tab:notation}%
	}
\end{table}%

\subsection{$Presentation$ $Attack$ Subset of FPAMS Database}
\label{ssec:PAD-datase}

The presentation artifact samples of the FPAMS database comprise $8$ artifacts, namely, from $2$ printed photographs, $4$ electronic displays, and $2$ face masks, acquired in a controlled environment. 
High-resolution $24$ MegaPixel color $Bonafide$ sample images corresponding to $145$ subjects collected using a DSLR (Model:D320) camera during $bonafide$ sample collection were used to generate print and electronic display attacks. 
\textbf{Print Artifacts:} Two artifacts were generated on high quality papers using two separate printers: Laser printer (Model: RICOH ATICO MP C4520) and InkJet printer (Model: HP Photosmart 5520). Using these two PAIs, we generated high-quality artifacts for the same $145$ $Bonafide$ samples, which were subsequently presented to a multispectral imaging sensor to introduce an attack. The samples were collected in a single session with $six$ sample images acquired for each artifact, which corresponds to $145$ subjects $\times$ $6$ samples $\times$ $9$ bands $\times$ $2$ Print Artifacts = $15660$ samples. 
\textbf{Electronic Display Artifacts:} For presenting this artifacts, we used $4$ electronic display that includes: (a) Apple iMAC 27-inch 5K Retina display, (b) Dell 27-inch 5K LED display, (c) Apple iPAD 9.7-inch Retina Display, and (d) Samsung Galaxy S8 5.8-inch display. 
High-quality digital images corresponding to the same $145$ subjects were presented independently using $4$ electronic display to acquire multispectral images. A total of $31320$ sample artifacts were acquired, corresponding to $145$ subjects $\times$ $6$ samples $\times$ $9$ bands $\times$ $4$ = $31320$ samples Electronic Display Artifacts. 
\textbf{Face Mask Artifacts:} To present this artifact species, we use rigid color and white face mask PAIs. 
Again, with the controlled lighting condition, we acquired a total of $1620$ artifacts that consisted of $18$ subjects $\times$ $5$ samples $\times$ $9$ bands $\times$ $2$ = $1620$ samples. 
Furthermore, for simplicity, notations are given for each artifact, as detailed in Table~\ref{tab:notation}. The acquired samples were then pre-processed to remove unwanted background information, normalized and cropped to $120\times120$ spatial resolution\cite{X.Zhu:face-detect-based-landmarks:CVPR:2012}.

\section{Multi-spectral Presentation Attack Detection (PAD): Combining Complementarary Inforamtion}
\label{sec:method}

In which complementary information from the multispectral imaging is combined using image fusion and score level fusion. Figure \ref{fig:PADProposed} illustrates the propose framework in which the spectral images are combined independently using image fusion and score level fusion. 
%%%%%%%% Update below ro have two sub section (a) Image Fusion (b) Score fusion
%This section of paper present the multi-spectral presentation attack detection framework of our work. The work is based on evaluating unseen training and testing samples towards exploring the potential of multi-spectral imaging sensor for PAD. Therefore, based on the FPAMS database, we learn the feature corresponding to artifacts during model training and testing is performed using unseen artifacts to present the robustness of employing multi-spectral imaging for PAD.  

Let the spectral band images represented by $M_\lambda (p,q)$ 
%%-------------------%%
\begin{equation}
	M_\lambda (p,q) =\left \{  M_1(p,q), M_2(p,q),\ldots,M_9(p,q)\right \}
	\label{eqn:spectra}
\end{equation}
%%-------------------%%  
where $\lambda$ indicates the spectral band images correspond to nine narrow bands, $(p,q)$ represents size of image i.e. $120\times120$ spatial resolution.

\textbf{Image fusion:} In this approach, we employed wavelet averaging fusion to combine the complementary spatial and spectral information. In general, we obtain first seven wavelet coefficients that comprises of a approximation, two vertical, two horizontal, and two diagonal coefficients using 2-level Descrite Wavelet Transform (2-DWT) as %$C_{\lambda} = \left \{ A_{\lambda}, V_{\lambda}, V^{'}_{\lambda},H_{\lambda}, H^{'}_{\lambda}, D_{\lambda}, D^{'}_{\lambda}  \right \}$,  %
The representation of these wavelet coefficients can be seen from Equation~\ref{eqn:DCT-coeff}.   
%%-------------------%%
\begin{equation}
	C_{\lambda} = \left \{ A_{\lambda}, V_{\lambda}, V^{'}_{\lambda},H_{\lambda}, H^{'}_{\lambda}, D_{\lambda}, D^{'}_{\lambda}  \right \} 
	\label{eqn:DCT-coeff}
\end{equation}
%%-------------------%%
where, approximation coefficient is indicated as $A_{\lambda}$, two vertical coefficient as $(V_{\lambda}, V^{'}_{\lambda})$, two horizontal coefficient as $(H_{\lambda}, H^{'}_{\lambda})$ and two diagonal coefficient as $(D_{\lambda}, D^{'}_{\lambda})$
Further, to fuse these seven coefficients, a weighted summation is performed using Equation~\ref{eqn:DCT-fus} as follows:
%%-------------------%%
\begin{equation}
	W_{fus} = \left \|  \omega_1 C_{1} + \omega_2 C_{2} + \omega_2 C_{2} + \ldots + \omega_9 C_{9} \right \|
	\label{eqn:DCT-fus}
\end{equation}
%%-------------------%%
Further, applying inverse transform on the fused coefficients to obtained final fused image used for Local Binary Pattern (LBP) features extraction followed by SVM classifier \cite{1717463}.  

\textbf{Score Fusion:} To avail the benefits of employing complementary information from each spectral bands we perform score fusion of individual spectral bands for PAD. Essentially, to leverage the discriminative spatial information across individual band, we engaged LBP texture descriptor, well proven method for local and global feature extraction. Not only does it extract the relevant features, but also reduce the dimension without compromising the performance. For instance, in this work $3\times3$ window size for LBP presents a feature vector of size $1\times256$ for each band in comparison to $120 \times 120$ spatial dimension. Let the feature extraction after performing LBP on Equation~\ref{eqn:spectra} be represented as:% and then the combined histogram features processed for PAD.        
%Let the spectral band images represented by $M_\lambda (p,q)$ 
%%-------------------%%
\begin{equation}
	\varphi _\lambda =\left \{  \varphi _1, \varphi _2,\ldots,\varphi _9\right \}
	\label{eqn:spectra-lbp}
\end{equation}
%%-------------------%%          
where $\varphi _\lambda \in \mathbb{R}^{1\times256}$ feature vector corresponding to individual spectral band and each having dimension of $1\times256$. Extracted feature vectors from individual spectral band were then processed independently using SVM classifier to obtain the prediction scores, which we further combined using simple sum rule to demonstrated our second approach of PAD. Equation~\ref{eqn:score-fuse} represent the score fusion to obtain final score.
%%------------------%%
\begin{equation}
	\Omega   = \left \|\omega _{(\lambda=1)} + \omega _{(\lambda=2)} +\ldots+\omega _{(\lambda=9)} \right \| 
	\label{eqn:score-fuse}
\end{equation}    
%%------------------%%  
where $\omega_{(\lambda=1,2,\ldots,9)}$ are the predicted scores from the classifier corresponding to individual spectral band and $\Omega$ represent the final output scores used for the performance analysis after employing sum rule to combine the scores.

\section{Experiments and Results}
\label{sec:results}

To present experimental evaluations, we perform extensive analysis of PAD on the Face Presentation Attack Multispectral (FPAMS) database. Referring to FPAMS, which consists of bonafide and eight artifact species from three different PAIs, we present an experimental evaluation protocol that comprises the training, development, and testing sets. For training partition we allocate $2300$ samples ($500$ Bonafide + $300$ $\times$ ($2$ Print Artifact + $4$ Display Artifact) PAIs), for development partition we allocate $1440$ samples ($300$ Bonafide + $180$ $\times$($2$ Print Artifact + $4$ Display Artifact) + $30$ $\times$ ($2$ Mask Artifact)) and testing set comprises of $2300$ samples ($650$ Bonafide + $390$ $\times$($2$ Print Artifact + $4$ Display Artifact) + $120$ $\times$ ($2$ Mask Artifact)). The data partition was disjoint and did not involve any overlap to avoid bias in the experimental evaluation. The development partition set is allocated mainly to compute the threshold value for $Bonafide$ and artifact species for the final evaluation with the testing set. To analyze the performance of the PAD algorithm, we used samples corresponding to two different PAIs in the training set and samples of other PAI in the testing set, which were not used in the training set. For instance, training with all the samples belongs to Display Artifacts and Print Artifacts, whereas the testing set consisted of Mask Artifact 1. The purpose of this study is to present an extensive evaluation of PAD and to explore the potential of multispectral imaging sensors on unseen artifact species. 
      %Bonafide samples and $180$ presentation artifacts each from the subsets of Print and Display PAIs, and $30$ face mask artifacts. Finally, Testing set set   As This section presents the experimental evaluation protocol and results 
\begin{table}[ht!]
	\centering
	\caption{Evaluation of PAD using score fusion approach}
	\resizebox{0.48\textwidth}{!}{
		\begin{tabular}{|c|c|c|c|c|c|}
			\hline
			\multirow{3}[6]{*}{\textbf{Training PAIs}} & \multirow{3}[6]{*}{\textbf{Testing PAI}} & \textbf{Development Set} & \multicolumn{3}{c|}{\textbf{Testing Set}} \bigstrut\\
			\cline{3-6}       &       & \multirow{2}[4]{*}{\textbf{D-EER}} & \multirow{2}[4]{*}{\textbf{D-EER}} & \multicolumn{2}{c|}{\textbf{BPCER @ APCER =}} \bigstrut\\
			\cline{5-6}       &       &       &       & \textbf{5\%} & \textbf{10\%} \bigstrut\\
			\hline
			All Display PAIs and all face mask PAIs & Print Artefact 1 & 3.33$\pm$1.37 & 3.09$\pm$1.08 & 2.03$\pm$1.99 & 0.66$\pm$0.67 \bigstrut\\
			\hline
			All Display PAIs and all face mask PAIs & Print Artefact 2 & 0.54$\pm$0.49 & 0.53$\pm$0.67 & 0.06$\pm$0.14 & 0.015$\pm$0.04 \bigstrut\\
			\hline \hline
			All Display PAIs and all print PAIs & Mask Artefact  1 & 0.00 $\pm$0.00      & 0.00 $\pm$0.00    & 0.00 $\pm$0.00      & 0.00 $\pm$0.00  \bigstrut\\
			\hline
			All Display PAIs and all print PAIs & Mask Artefact  2 & 0.00 $\pm$0.00     & 0.02$\pm$0.07 & 0.00 $\pm$0.00      & 0.00 $\pm$0.00  \bigstrut\\
			\hline \hline
			All face Mask PAIs and all print PAIs & Display Artefact 1 & 8.19$\pm$2.38 & 6.89$\pm$0.62 & 14.95$\pm$6.34 & 7.24$\pm$2.79 \bigstrut\\
			\hline
			All face Mask PAIs and all print PAIs & Display Artefact 2 & 30.42$\pm$2.36 & 29.40$\pm$2.63 & 71.77$\pm$7.77 & 57.50$\pm$7.52 \bigstrut\\
			\hline
			All face Mask PAIs and all print PAIs & Display Artefact 3 & 10.57$\pm$2.12 & 9.02$\pm$0.66 & 19.36$\pm$5.11 & 11.01$\pm$3.71 \bigstrut\\
			\hline
			All face Mask PAIs and all print PAIs & Display Artefact 4 & 0.43$\pm$0.49 & 0.58$\pm$0.13 & 0.04$\pm$0.07 & 0.00 $\pm$0.00  \bigstrut\\
			\hline
		\end{tabular}%
		\label{tab:textureDiscriptor}%
	}
\end{table}%
\begin{table}[ht!]
	\centering
	\caption{Evaluation of PAD using spectral Image Fusion Method}
	\resizebox{0.48\textwidth}{!}{
		% Table generated by Excel2LaTeX from sheet 'Sheet2'
		\begin{tabular}{|p{18em}|c|c|c|c|c|}
			\hline
			\multicolumn{1}{|c|}{\multirow{3}[6]{*}{\textbf{Training PAIs}}} & \multirow{3}[6]{*}{\textbf{Testing PAI}} & \textbf{Development Set} & \multicolumn{3}{c|}{\textbf{Testing Set}} \bigstrut\\
			\cline{3-6} \multicolumn{1}{|c|}{} &       & \multirow{2}[4]{*}{\textbf{D-EER}} & \multirow{2}[4]{*}{\textbf{D-EER}} & \multicolumn{2}{c|}{\textbf{BPCER @ APCER =}} \bigstrut\\
			\cline{5-6} \multicolumn{1}{|c|}{} &       &       &       & \textbf{5\%} & \textbf{10\%} \bigstrut\\
			\hline
			All Display PAIs and all face mask PAIs & Print Artefact 1 &   50    &    50   & 100   & 100 \bigstrut\\
			\hline
			All Display PAIs and all face mask PAIs & Print Artefact 2 &  50     &   50    & 100   & 100 \bigstrut\\
			\hline \hline
			All Display PAIs and all print PAIs & Mask Artefact  1 & 12.67+3.71 & 14.79+2.42 & 47.46+17.16 & 32.32+15.10 \bigstrut\\
			\hline
			All Display PAIs and all print PAIs & Mask Artefact  2 & 10.90+5.62 & 9.65+4.63 & 21.55+12.77 & 16.38+11.56 \bigstrut\\
			\hline \hline
			All face Mask PAIs and all print PAIs & Display Artefact 1 & 32.43+21.48 & 31.90+20.80 & 67.33+40.56 & 61.20+43.15 \bigstrut\\
			\hline
			All face Mask PAIs and all print PAIs & Display Artefact 2 & 29.36+21.20 & 29.44+21.70 & 65.60+42.96 & 59.29+43.38 \bigstrut\\
			\hline
			All face Mask PAIs and all print PAIs & Display Artefact 3 & 28.30+20.19 & 28.81+22.00 & 63.53+43.17 & 55.58+43.63 \bigstrut\\
			\hline
			All face Mask PAIs and all print PAIs & Display Artefact 4 & 28.20+20.16 & 30.11+22.92 & 65.81+42.38 & 58.10+43.52 \bigstrut\\
			\hline
		\end{tabular}%
		\label{tab:image-fuse}%
	}
\end{table}%
We present the results of the PAD algorithm using performance metrics such as: (a) Attack Presentation Classification Error Rate (APCER $\%$) - percentage error of sample presentation attack from PAIs classified as bonafide presentation, Bonafide Presentation Classification Error Rate (BPCER $\%$) -  percentage error of bonafide samples classified as presentation attack. Based on these performance metrics, we present the performance of BPCER when the operating point with APCER = $5\%$ and $10\%$ and Detection - Equal Error Rate (D-EER) when APCER equals BPCER on the development set as well as with the testing set.

\begin{figure}[htbp!]
	\centering
	\includegraphics[width = 0.9\linewidth]{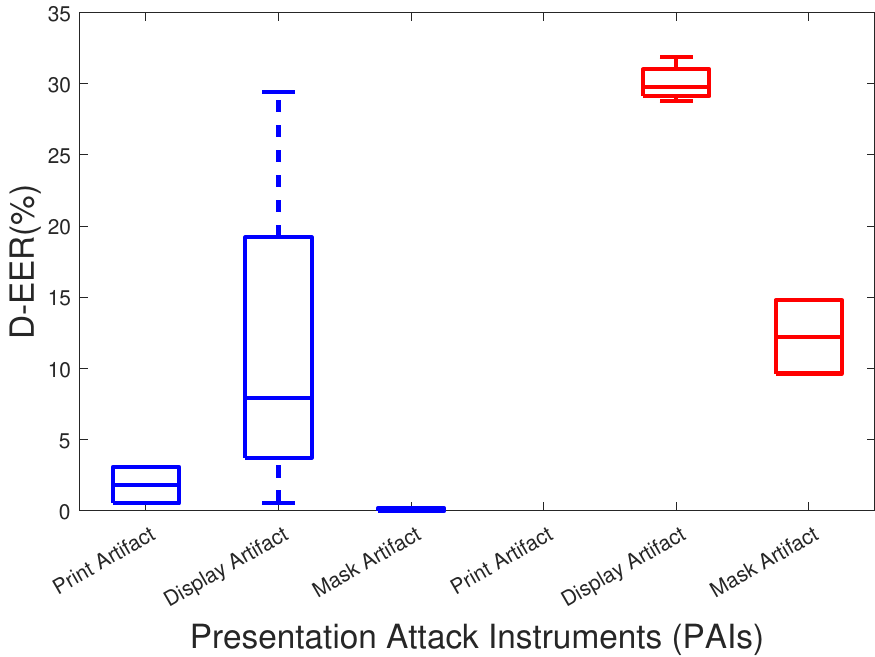}
	\caption{Mean and variance plot across PAIs: Blue color indicate score fusion and Red Color indicate image fusion}
	\label{fig:PAD-boxplot}
\end{figure}

In this study, to leverage the complementary details across individual spectral bands, we present the performance of PAD based on score fusion and image fusion methods. Table~\ref{tab:textureDiscriptor} and ~\ref{tab:image-fuse} represents the quantitative experimental evaluation results computed by employing leave one out approach on training, development and testing set partition with no overlap in each of the subsets. Figure~\ref{fig:PAD-boxplot} illustrates the mean-variance plot showing the performance comparison across individual artifact species and the two different methods used in this study. From the obtained results, PAD based on the score fusion outperformed the image fusion algorithms independently across all artifacts. This implies that there is a significant amount of distinction between the spectral reflectance properties of skin and non-skin (artifacts from PAIs) owing to their better classification accuracy. It is further evident from Figure~\ref{fig:PAD-boxplot} of the mean-variance plot illustrating the lower D-EER of the score fusion compared with the image fusion approach.   
%---------------%
%--------------%
%Among the two different methods applied in this work, LBP based texture descriptor method performs better compared to the image fusion algorithm. 
Specifically, $0.00\%$ D-EER is obtained along with $0.00\%$ BPCER at $5\%$ and $10\%$ APCER error, whereas the performance of the spectral image fusion algorithms degrades, as can be seen from the tabular results (Table~\ref{tab:image-fuse}). 
Although the lowest D-EER is obtained for the score fusion algorithm, the performance of this method is observed to be better across print and face mask artifacts and partially in display artifacts (i.e., only for two display artifact species). %, performance analysis is uniformly better across all the artifacts based on spectral signature approach. 
The reason for the slightly poor results could be the very low signal-to-noise ratio (SNR) or absence of information across bands such as $770nm$, $830nm$, $950nm$,and $1000nm$ (electronic display does not emit illumination in this wavelength region). 
However, image fusion forms a composite image, and a low signal-to-noise ratio in these bands certainly contributes to the worst performance compared to the individual band score fusion, as evident from Table~\ref{tab:image-fuse} and Figure~\ref{fig:PAD-boxplot}.    
%However, in comparison to texture descriptor, image fusion method perform worst  even results can be observed from Table~\ref{tab:image-fuse}Due to the formation of composite image using image fusion algorithm on these low SNR images may results in poor performance of this algorithm. 
Furthermore, both algorithms have shown better classification performance with print and face mask artifacts, while the reason for low performance with display artifacts, as stated above, is the low SNR; hence, the performance with display artifacts is not observed consistently well across the four different artifact species of display attack.     
To summarize, the performance of PAD based on image fusion and score fusion is reasonable and signifies the use of spectral properties of multispectral imaging to improve presentation attack detection accuracy.

\section{Conclusion}
\label{sec:conclusion}

Vulnerability of face recognition systems has been challenged by various presentation attack instruments. With a substantial amount of work in detecting presentation artifacts in the visible spectrum domain, multi-spectral imaging sensors have gained significant attention in this direction for their robust performance. In this work, we present a PAD based on a multispectral imaging sensor to explore its inherent differential illumination properties across presentation attack instruments in comparison with the bonafide. We present a performance analysis using a newly introduced FPAMS database consisting of eight different artifacts, including two print, four electronic displays, and two face mask artifacts.  The results obtained from $61650$ sample spectral band images comprised bonafide and artifact data collected in nine narrow bands across the VIS and NIR ranges. The evaluation results obtained using the two different methods include image fusion and score fusion. Based on the obtained results, best result of BPCER=$0\%$ at APCER=$5\%$ and $10\%$ signifies the superiority of multispectral imaging in detecting presentation artifacts.

%\clearpage

%\bibliographystyle{IEEEbib}
%\balance
\scriptsize
%\small
\bibliographystyle{IEEEtran}
\bibliography{reference}

\begin{thebibliography}{10}
\providecommand{\url}[1]{#1}
\csname url@samestyle\endcsname
\providecommand{\newblock}{\relax}
\providecommand{\bibinfo}[2]{#2}
\providecommand{\BIBentrySTDinterwordspacing}{\spaceskip=0pt\relax}
\providecommand{\BIBentryALTinterwordstretchfactor}{4}
\providecommand{\BIBentryALTinterwordspacing}{\spaceskip=\fontdimen2\font plus
\BIBentryALTinterwordstretchfactor\fontdimen3\font minus
  \fontdimen4\font\relax}
\providecommand{\BIBforeignlanguage}[2]{{%
\expandafter\ifx\csname l@#1\endcsname\relax
\typeout{** WARNING: IEEEtran.bst: No hyphenation pattern has been}%
\typeout{** loaded for the language `#1'. Using the pattern for}%
\typeout{** the default language instead.}%
\else
\language=\csname l@#1\endcsname
\fi
#2}}
\providecommand{\BIBdecl}{\relax}
\BIBdecl

\bibitem{article}
D.~Sharma and A.~Selwal, ``A survey on face presentation attack detection
  mechanisms: hitherto and future perspectives,'' \emph{Multimedia Systems},
  vol.~29, pp. 1--51, 03 2023.

\bibitem{8714076}
A.~George, Z.~Mostaani, D.~Geissenbuhler, O.~Nikisins, A.~Anjos, and S.~Marcel,
  ``Biometric face presentation attack detection with multi-channel
  convolutional neural network,'' \emph{IEEE Transactions on Information
  Forensics and Security}, vol.~15, pp. 42--55, 2020.

\bibitem{ISO-IEC-JTC-SC37-2013}
{ISO/IEC JTC1 SC37 Biometrics}, \emph{ISO/IEC DIS 30107-3. Biometrics-
  Presentation Attack Detection - Part 3}, Testing and Reporting, International
  Organization for Standardization and International Electrotechnical
  Committee, August, 2016.

\bibitem{8014830}
R.~Raghavendra, K.~B. Raja, S.~Venkatesh, and C.~Busch, ``Face presentation
  attack detection by exploring spectral signatures,'' in \emph{2017 IEEE
  Conference on Computer Vision and Pattern Recognition Workshops (CVPRW)},
  2017, pp. 672--679.

\bibitem{george2022comprehensive}
A.~George, D.~Geissbuhler, and S.~Marcel, ``A comprehensive evaluation on
  multi-channel biometric face presentation attack detection,'' 2022.

\bibitem{Bhattacharjee2019}
S.~Bhattacharjee, A.~Mohammadi, A.~Anjos, and S.~Marcel, \emph{Recent Advances
  in Face Presentation Attack Detection}.\hskip 1em plus 0.5em minus
  0.4em\relax Cham: Springer International Publishing, 2019, pp. 207--228.

\bibitem{iet:content-journals}
A.~Costa‐Pazo, \emph{IET Biometrics}, vol.~10, pp. 408--429(21), July 2021.

\bibitem{7738245}
N.~Vetrekar, R.~Raghavendra, and R.~Gad, ``Low-cost multi-spectral face imaging
  for robust face recognition,'' in \emph{2016 IEEE International Conference on
  Imaging Systems and Techniques (IST)}, 2016, pp. 324--329.

\bibitem{X.Zhu:face-detect-based-landmarks:CVPR:2012}
X.~Zhu and D.~Ramanan, ``Face detection, pose estimation, and landmark
  localization in the wild,'' in \emph{IEEE Conference on Computer Vision and
  Pattern Recognition}, June 2012, pp. 2879--2886, software available at
  \url{https://www.ics.uci.edu/~xzhu/face/}.

\bibitem{1717463}
T.~Ahonen, A.~Hadid, and M.~Pietikainen, ``Face description with local binary
  patterns: Application to face recognition,'' \emph{IEEE Transactions on
  Pattern Analysis and Machine Intelligence}, vol.~28, no.~12, pp. 2037--2041,
  2006.

\end{thebibliography}

%\begin{thebibliography}{00}
%\bibitem{b1} G. Eason, B. Noble, and I. N. Sneddon, ``On certain integrals of Lipschitz-Hankel type involving products of Bessel functions,'' Phil. Trans. Roy. Soc. London, vol. A247, pp. 529--551, April 1955.
%\bibitem{b2} J. Clerk Maxwell, A Treatise on Electricity and Magnetism, 3rd ed., vol. 2. Oxford: Clarendon, 1892, pp.68--73.
%\bibitem{b3} I. S. Jacobs and C. P. Bean, ``Fine particles, thin films and exchange anisotropy,'' in Magnetism, vol. III, G. T. Rado and H. Suhl, Eds. New York: Academic, 1963, pp. 271--350.
%\bibitem{b4} K. Elissa, ``Title of paper if known,'' unpublished.
%\bibitem{b5} R. Nicole, ``Title of paper with only first word capitalized,'' J. Name Stand. Abbrev., in press.
%\bibitem{b6} Y. Yorozu, M. Hirano, K. Oka, and Y. Tagawa, ``Electron spectroscopy studies on magneto-optical media and plastic substrate interface,'' IEEE Transl. J. Magn. Japan, vol. 2, pp. 740--741, August 1987 [Digests 9th Annual Conf. Magnetics Japan, p. 301, 1982].
%\bibitem{b7} M. Young, The Technical Writer's Handbook. Mill Valley, CA: University Science, 1989.
%\end{thebibliography}

\end{document}